# What sentiment analysis can't see

*Measuring whether customers were helped, and what went wrong, across 70,000 support conversations*

Jason Potteiger

Dimension Labs

June 2026

## Abstract

Most companies read their customer support data at scale using sentiment analysis, which measures how customers sound rather than whether they were satisfied with the result. We tested a richer alternative on 70,450 support conversations from a leading online fundraising platform: alongside tone, we used GPT-5.4 to estimate each customer's satisfaction and to flag whether they reported a concrete problem, then validated all three readings against the 1-to-5 ratings customers left on the conversations they rated. The satisfaction estimate tracked those ratings far better than sentiment did, correlating at 0.47 against 0.36 and flagging unhappy customers with far fewer false alarms. The structured read also sees what sentiment cannot: tone and satisfaction disagree in 44% of conversations, a single "Neutral" label hides everything from quietly satisfied customers to ones who quietly gave up, and the largest group of all is "tolerated friction," customers who are satisfied but still reporting a fixable problem, a standing issue that no sentiment-based dashboard can surface. The broader finding is that LLM-based annotation can capture far more than the tonality of a customer’s language, offering strong potential for new business metrics grounded instead in the customer’s state (whether they were satisfied) and the cause of their problem extracted from the raw textual data of interactions and feedback.

## 1. Introduction

A customer writes a few civil messages about a refund that will not go through, thanks the agent for the help, and then rates the interaction a 1, with the remark that it was "too convoluted." Read as tone, the conversation is unremarkable; read as a verdict (the self-reported rating), it failed. The two readings describe the same customer and the same exchange, and they point in opposite directions. This is the situation the paper investigates. When an organization summarizes its support operation with sentiment analysis, it is reading the tone of customers' language and treating that tone as a stand-in for how customers judge the service or the severity of their issue. We ask what that stand-in actually measures, whether a model's structured read of the customer's state and the operational cause tracks the customer's own verdict more faithfully, and what this method recovers that sentiment cannot see.

The data are 70,450 customer-support conversations from a leading online fundraising platform, exchanged through Intercom/Fin between November 2025 and April 2026 and passing structural gates for genuine, non-trivial exchanges. We annotated each conversation with four independent prompt functions run with GPT-5.4, reading the same text but never sharing a prompt, so that agreement or disagreement among their outputs is a property of the constructs rather than of shared anchoring. Three readings concern us here: the customer's tone (sentiment), a holistic 1-to-5 estimate of the customer's satisfaction (predicted satisfaction), and a flag for whether the customer reported a concrete problem (issue presence). The model received only the conversation text, not wait times, ticket metadata, or the customer's rating. On the 5.55% of conversations that carried a customer-provided 1-to-5 rating (N = 3,911), we validated each reading against that rating, leading with metrics that survive the skew toward favorable ratings.

Predicted satisfaction is the more valid of the two readings by a clear margin. It correlates with the customer's rating at 0.47 against sentiment's 0.36, a difference whose interval excludes zero; it identifies low ratings with roughly twice the precision (0.51 against 0.29); and its bands map monotonically onto observed ratings, **which makes it usable as a proxy on the conversations that are never rated**. Sentiment is not worthless, but it is the weaker instrument, and it pays for its errors at the unfavorable end of the scale. The more consequential result appears when the two readings are compared with each other rather than with the rating. They disagree in 44% of conversations, and the disagreement is not noise: it concentrates almost entirely in conversations whose tone is neutral or negative but whose predicted satisfaction is high. That asymmetry raises the questions the rest of the paper answers: what a structured read of state[1] and cause[2] recovers that sentiment analysis compresses, why the two part company so often, and what the gap costs an organization that monitors only tone.

What follows has direct consequences for anyone monitoring support at scale. If tone tracks how a customer sounds rather than whether they were helped, and Section 4 shows sentiment's negative flag is right less than a third of the time it fires, then the sentiment dashboards most teams rely on are measuring the wrong thing. Reading the customer's state and the operational cause instead changes what a team can see: it tells apart the customer who is quietly satisfied from the one who is quietly giving up, it surfaces the large backlog of fixable problems that satisfied customers absorb without complaint, and, because it runs on every conversation rather than the rated few, it exposes risk that ratings-based monitoring never reaches. We establish the measurement result first, then trace what it recovers.

[1] The customer's state is how the interaction left them, satisfied or not, as estimated here by predicted satisfaction.
[2] The operational cause is the concrete problem behind the contact, if any, as flagged here by issue presence.

## 2. Related research

The conventional way to summarize unstructured customer feedback at scale is sentiment analysis, and the field that built it has spent two decades moving away from the version still running on most dashboards. Pang and Lee (2008) and Liu (2015) codified the task as classifying the polarity of a document or sentence as positive, negative, or neutral, and that document-level label remains the default operational metric in customer-experience work. But the same canon treated polarity as a starting point rather than a destination. Hu and Liu (2004) reframed opinion mining as the extraction of specific product features and the sentiment attached to each, and the SemEval aspect-based sentiment task (Pontiki et al., 2014) formalized opinions as structured objects defined by an entity, an aspect, and a stance toward it. The trajectory is unmistakable: from "is this text positive or negative" toward "what is the customer talking about, and what is their position on it." A representation built from the customer's state and the operational cause is the next step along that line, and document-level sentiment is the baseline against which it should be measured.

The deeper problem with polarity is not that it is coarse but that it was never validated as a measure of what it is used for. Puschmann and Powell (2018) document how sentiment scores came to be treated as a currency for consumer preference without anyone establishing that polarity measures preference, a construct claim naturalized through repetition rather than evidence. Mozetič et al. (2016) supply the empirical counterpart: across a large multilingual corpus, classifier performance is bounded by the agreement among human annotators, which means the ground truth the field trains and evaluates against is itself unstable, a ceiling on the measure that no better model can lift because the instability lives in the construct rather than the algorithm. And where polarity has been carried into a high-stakes enterprise domain it has failed concretely, with Loughran and McDonald (2011) finding that roughly three-quarters of the words a standard dictionary flags as negative are not negative in financial text, so an off-the-shelf scorer misreads the documents it is most often deployed on. Contextual models have since narrowed this particular failure: transformer-based classifiers and, more recently, LLMs read polarity in context far better than a fixed word list, rarely mistaking a term like "liability" for a negative signal. But the gain is in lexical accuracy, not construct validity; a more accurate reading of tone is still a reading of tone, not a measure of how the customer judged the interaction. This is the backdrop against which the present study's central measurement question matters. Showing that another measure aligns with customers' own ratings better than sentiment does is not a foregone conclusion dressed up as a finding; it is a test of a default whose validity has never been secure.

If sentiment is a flawed proxy, the alternative is to estimate the customer's state directly and hold it to the customer's own verdict. Two developments make that practical. First, even in the era of large language models, classifying polarity and extracting structured information remain distinct tasks with distinct performance profiles (Zhang et al., 2024), so a structured estimate is not simply sentiment computed by a larger model. Second, language models can produce scalar measurements of latent constructs from open-ended text (Licht, 2025) and can predict Likert-scale ratings from review text (Ryu and Yanaka, 2025), though scalar scoring carries its own failure modes (output bunching and directional bias) that only validation against a real criterion can detect. Prior work on this platform made the move directly: an LLM-predicted experience score, validated against fan-reported survey ratings, recovered the rating within a point most of the time and, more tellingly, diverged from it in a systematic rather than a random way (Hong, Potteiger, and Zapata, 2026). The present study applies the same discipline to customer support. It estimates satisfaction from conversation text with an independent classifier, validates that estimate against the customer's own rating, and then asks what the places where the two part company reveal.

That a text-derived estimate and a self-reported rating should diverge at all is what the psychology of self-report predicts, and the reason is worth stating before the evidence arrives. Schwarz (1999) shows that the format of a question determines what a respondent retrieves: an open-ended prompt pulls the salient, recent, and concrete, while a rating scale pulls a holistic evaluative summary. That summary draws heavily on the respondent's momentary affect rather than a careful tally of events (Schwarz and Clore, 1983), and it is weighted toward the emotional peak and the ending rather than the average of an experience (Kahneman et al., 1993; Kahneman, Wakker, and Sarin, 1997). The implication for any text-based measure is direct. This study, though, is not the survey setting that produced these findings, and the difference matters. In a survey, the two signals are two self-reports: a structured rating, which yields a **verdict**, and an open-ended answer, which is governed by **salience**, whatever was vivid enough to mention. Here there is only one self-report, the 1-to-5 rating the customer leaves at the end, and it is the verdict. The other signal is not the customer's open-ended answer but the **full transcript of the interaction**, every message on both sides. We are therefore not trying to reconstruct the rating from a customer's salient remarks, as one would when mining open-ended survey text; we are reading the whole picture, which holds far more than what the customer chose to highlight, and that is what makes a verdict-like judgment recoverable from text at all. The two readings we take of that transcript fall along the axis the survey literature describes. Sentiment, which registers the affective surface of the language, stays close to salience; a measure like predicted satisfaction is asked to judge how the whole interaction went, integrating the same full text into something closer to the verdict. Both readings see the same source, so what separates them is not what they see but what they are asked to do. We return to this distinction in Section 7 to interpret where tone and state diverge; here it establishes only that the divergence is to be expected, not treated as a defect in either reading.

**Terminology: the survey contrast, mapped to conversational data.** *In a survey the two signals are two self-reports; here only the verdict is self-reported, and the second signal is the full transcript, read two ways.*

| Concept | In survey research | In this study (conversation transcripts) |
|---|---|---|
| **Verdict** | the structured rating: a holistic judgment | the 1-to-5 rating the customer leaves at the end (our criterion) |
| **Salience** | the open-ended answer: whatever was vivid enough to report | the affective surface of the language, which sentiment reads |
| **The full record** | (a survey has no equivalent) | the entire transcript, which predicted satisfaction integrates to approximate the verdict |

## 3. Data, annotation, and validation design

Every measure in this study is recovered from a single artifact: the text of a support conversation. The customer's tone, the presence of a concrete problem, and the customer's overall satisfaction are not separate data streams collected by separate instruments. They are three readings of the same words, produced by three independent AI prompt functions. This is what makes the conversation a clean site for a construct question. When two readings of the same text disagree, the disagreement cannot be blamed on different inputs, different time windows, or different populations. It is a property of what each reading is measuring.

**The corpus.** The data are customer support conversations from an all-in-one fundraising platform, exchanged through Intercom/Fin between November 2025 and April 2026. The platform is a leader in the online donation space that processes billions of dollars in donations annually; in the donor management software category, it maintains exceptionally high ratings, driven by praise for its modern user interface and customer support. Incoming messages are from platform users; outgoing messages are from support agents. We deduplicated each annotation table and the account metadata table to one record per conversation and applied three structural gates drawn from a context-level classifier: the exchange was a genuine two-party conversation (`g2_conversation_true`), it was not a fragment (`g2_full_conversation_true`), and it contained at least three messages (`g2_message_count >= 3`). The three-message floor removes one- and two-turn interactions that carry too little text to read while preserving the substantive conversations. After gating, the analyzable corpus is 70,450 conversations, drawn from 75,066 deduplicated annotated conversations. The corpus is dominated by medium-length exchanges: roughly half are eight to twelve messages, and most of the remainder run longer.

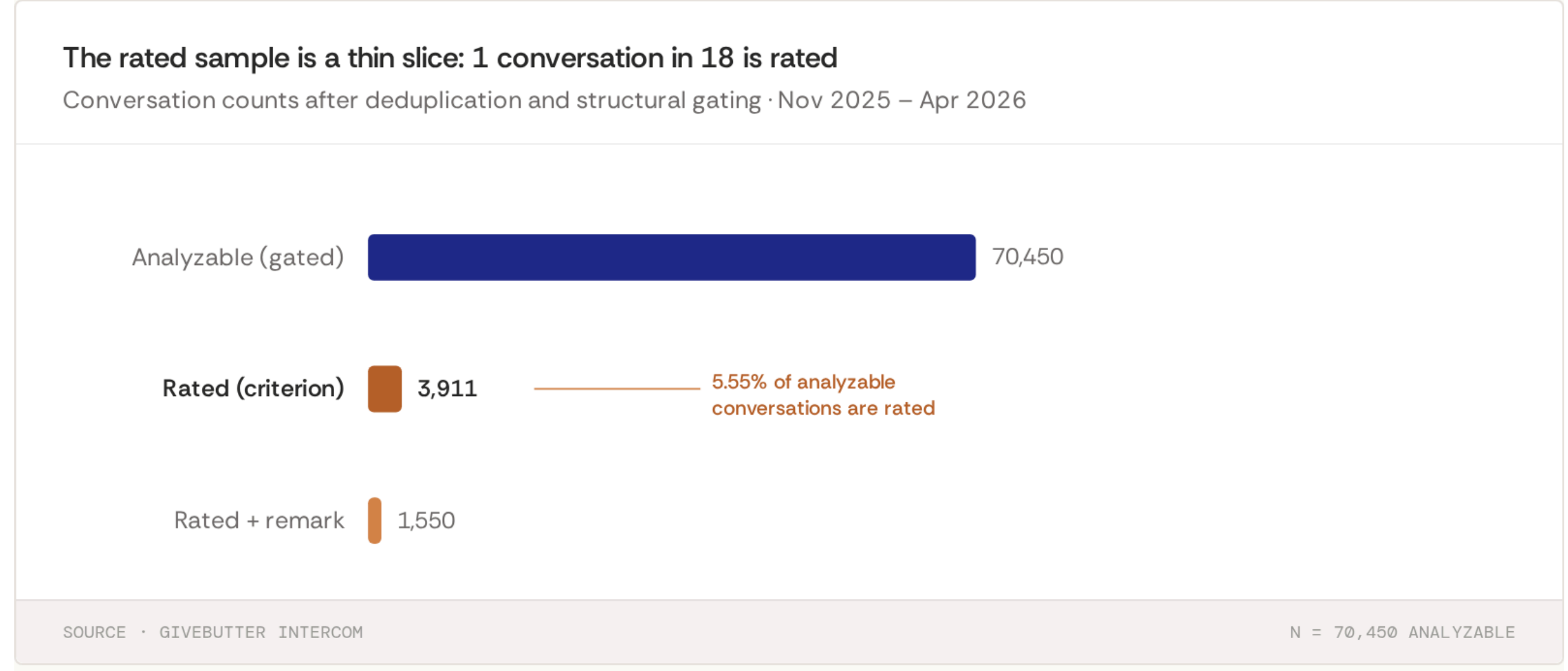

*Figure 1.*

**The criterion.** A small share of these conversations carry the measure we validate against. After an interaction, a customer may leave a 1-to-5 support rating (`rating_value`), and may attach a free-text remark (`rating_remark`). Ratings are present on 5.55% of analyzable conversations (N = 3,911) and remarks on 2.20% (N = 1,550). The rating is the closest thing the data contain to the customer's own verdict on the interaction, and it is the primary criterion throughout. It is also skewed: of the 3,911 rated conversations, 2,755 are 5s and only 355 are low (a 1, 2, or 3), a low-rating base rate of 9.08%. A criterion this favorable rewards any measure that simply predicts "good" most of the time, so raw agreement is uninformative on its own and we report it only beside a majority-class baseline. The claims that follow lead with metrics that survive the skew.

**Recognition, not generation.** The annotation ran on the Dimension Labs language data platform, which organizes unstructured customer text into a common schema and labels it with large language models for agentic analysis. The labeling is a recognition task. Each classifier read one conversation's text and returned a structured set of fields describing information already present in that text; it did not generate new content, infer beyond the words, or carry memory from one conversation to the next. Four classifiers, implemented as independent prompt functions and run with GPT-5.4, annotated the same conversations. A context function established the structural gates and an agent-handling score (`g2_success_rating`). A sentiment function returned the customer's tone. An issue function detected whether the customer reported a concrete problem. A satisfaction function estimated the customer's overall satisfaction. Because the four functions are separate prompts, any agreement or disagreement among their outputs reflects distinct measurements of the same conversation rather than shared anchoring inside a single prompt. This independence is the design feature that makes the divergence analysis in Sections 5 through 7 interpretable.

**The three readings and the criterion.** The study sets one conventional measure against two structured ones. The conventional measure is **sentiment** (`g2_user_sentiment`), a three-category label (Positive, Neutral, or Negative) for the affective tone of the customer's language. The structured measures are **predicted satisfaction** (`g2_predicted_satisfaction`), a holistic 1-to-5 estimate of the customer's satisfaction with the interaction, and **issue presence** (`g2_issue_mention`), a Boolean flag for whether the customer reported a specific problem. Predicted satisfaction is holistic and text-only by construction: it judges the conversation as a whole rather than scoring any single moment or feature, and it reads only what the customer's messages reveal, not wait times or ticket metadata. Table 1 lists each field, whether it is self-reported or model-annotated, and its definition.

**Table 1. Measures and criterion used in the study.** *The criterion is self-reported by the customer; the three readings are recovered from conversation text by independent GPT-5.4 classifiers.*

| Attribute | Source | Definition |
|---|---|---|
| `rating_value` | Self-reported | Customer's 1–5 rating of the support interaction (criterion). |
| `rating_remark` | Self-reported | Free-text comment left with a rating, where present. |
| `g2_user_sentiment` | LLM-annotated | Tone of the customer's messages: Positive, Neutral, or Negative. |
| `g2_predicted_satisfaction` | LLM-annotated | Holistic 1–5 estimate of the customer's satisfaction, from text alone. |
| `g2_predicted_satisfaction_confidence` | LLM-annotated | How diagnostic the text is for the assigned scale point: High, Medium, or Low. |
| `g2_issue_mention` | LLM-annotated | Whether the customer reported a specific problem (True/False). |
| `g2_issue_reason` | LLM-annotated | Short free-text summary of the reported issue, where present. |
| `g2_success_rating` | LLM-annotated | Context-function estimate of how well the agent handled the request (0–10). |

**What the confidence label measures.** The satisfaction function attaches a confidence label to every rating, and it is easy to misread. It does not report the model's introspective certainty, and it does not report how strong the customer's satisfaction or dissatisfaction is. It reports how clearly the customer's words point to one specific scale point rather than another, whether the language is decisive enough to separate, say, a 4 from a 5. A High label means the customer's language maps cleanly to one scale point; a Low label means the text is too thin or too mixed to commit to a direction. This matters because confidence concentrates at predictable places: High at the endpoints, where customer language is most diagnostic, and Low at the midpoint, where a 3 often reflects insufficient evidence rather than genuine ambivalence. We use the label in Section 5 to test whether the divergence between tone and state is merely a low-confidence artifact, and the distinction between a confident 3 and an uncertain 3 is what makes that test meaningful.

**Validation metrics.** Because the criterion is thin and favorable, the validation in Section 4 leads with measures that do not reward majority-class prediction. For the full 1-to-5 scale we report the rank correlation between each reading and `rating_value`, and we test the predicted-satisfaction-versus-sentiment comparison as a confidence interval on the *difference* between their correlations rather than as two separate point estimates. For detecting the conversations that matter operationally (the low ratings), we report precision, recall, specificity, balanced accuracy, and the Matthews correlation coefficient, all of which penalize a measure that earns its accuracy by calling everything favorable. We report a collapsed favorable-versus-unfavorable agreement beside the majority-class baseline so that lift is explicit. Finally, we examine how predicted-satisfaction bands map onto observed ratings, which turns the validated score into a calibration map usable on the conversations that are never rated.

## 4. Predicted satisfaction tracks the customer's rating; sentiment trails it

If predicted satisfaction and sentiment were interchangeable summaries of a support conversation, they would track the customer's own rating about equally well. They do not, and the gap is the foundation everything else in this paper rests on. Predicted satisfaction correlates with `rating_value` at 0.47; sentiment correlates at 0.36. Tested as a difference rather than as two point estimates, the gap runs from roughly +0.09 to +0.18 and excludes zero, so the advantage is not an artifact of where we drew the sample. Neither measure is a perfect proxy for the rating, but the structured estimate is the better of the two by a margin that holds up under the test that was fixed in advance.

**Table 2. Predicted satisfaction outperforms sentiment against the self-reported rating on every skew-robust metric.** *Rated criterion sample, N = 3,911. Low ratings are 1–3 (base rate 9.08%).*

| Metric | `g2_predicted_satisfaction` | `g2_user_sentiment` |
|---|---|---|
| Correlation with `rating_value` | 0.47 | 0.36 |
| Low-rating precision | 0.51 | 0.29 |
| Low-rating recall | 0.31 | 0.30 |
| Specificity | 0.97 | 0.93 |
| Balanced accuracy | 0.64 | 0.61 |
| Matthews correlation | 0.35 | 0.22 |
| Favorable/unfavorable accuracy | 91% | 74% |

**The advantage is in precision, not recall.** The two measures find low ratings at almost the same rate: predicted satisfaction recovers 31% of low-rated conversations, sentiment 30%. The difference is in how much noise each drags along. When predicted satisfaction flags a conversation as low, it is right about half the time (precision 0.51); when sentiment flags one, it is right about a quarter of the time (precision 0.29). For a team that uses the signal to decide which conversations to review, that is the difference that matters. A flag worth acting on is one that is usually correct, and sentiment's negative label is correct on this criterion less than a third of the time. The same pattern shows up in the summary measures that account for the skew: balanced accuracy is 0.64 versus 0.61, and the Matthews correlation, which is unforgiving of majority-class guessing, is 0.35 versus 0.22. One might expect a more accurate measure to win by catching more of the bad cases. Predicted satisfaction wins instead by raising fewer false alarms.

**Collapsed agreement, with the baseline in view.** Reduced to a two-way favorable-versus-unfavorable judgment, predicted satisfaction agrees with the customer 91% of the time and sentiment 74%. The 91% figure should be read against the majority-class baseline: because 91% of ratings are favorable, a measure that simply called every conversation favorable would also score near 91% on accuracy alone. What distinguishes predicted satisfaction is that it reaches that agreement while still discriminating at the unfavorable end, which the precision and balanced-accuracy figures confirm and which the always-favorable baseline cannot do. Sentiment, by contrast, sits well below the baseline, because its Negative label fires on conversations the customer went on to rate favorably. Tone disagrees with the verdict often enough to cost it accuracy that the trivial baseline keeps for free. This is the construct-validity gap of Section 2 (Puschmann and Powell, 2018; Loughran and McDonald, 2011) appearing as a number: a polarity label that reads the surface language is not a measure of how the customer judged the interaction, and the rated data show the cost of treating it as one.

**The score is calibrated enough to deploy.** The validation would be a curiosity if the predicted score did not map onto the rating in an orderly way, so we examined how each band behaves against observed ratings. The mapping is monotonic across the scale. Conversations the model scores a 5 carry an average customer rating of 4.7 and a low-rating share of about 3%; conversations it scores a 1 carry an average rating of 2.0 and a low-rating share near 86%. The midpoint is the least stable region, as expected: a predicted 3 spans average ratings from roughly 3.4 to 3.6 depending on confidence, and absorbs the conversations where the text did not commit to a direction. The predictions concentrate at the top of the scale, which partly reflects the genuine positive skew of resolved support interactions but is also consistent with the output bunching documented for LLM scalar scores (Licht, 2025); it is one reason we validate the bands against observed ratings rather than reading meaning into the raw distribution. But the ends of the scale are sharp, and that is what a monitoring proxy needs. The practical reading is a four-level risk map: a 5 is very low risk, a 4 is low risk worth watching for issue density, a 3 is genuinely uncertain, and a 1 or 2 is elevated concern. Section 6 uses this map to extend the measure to the conversations that are never rated.

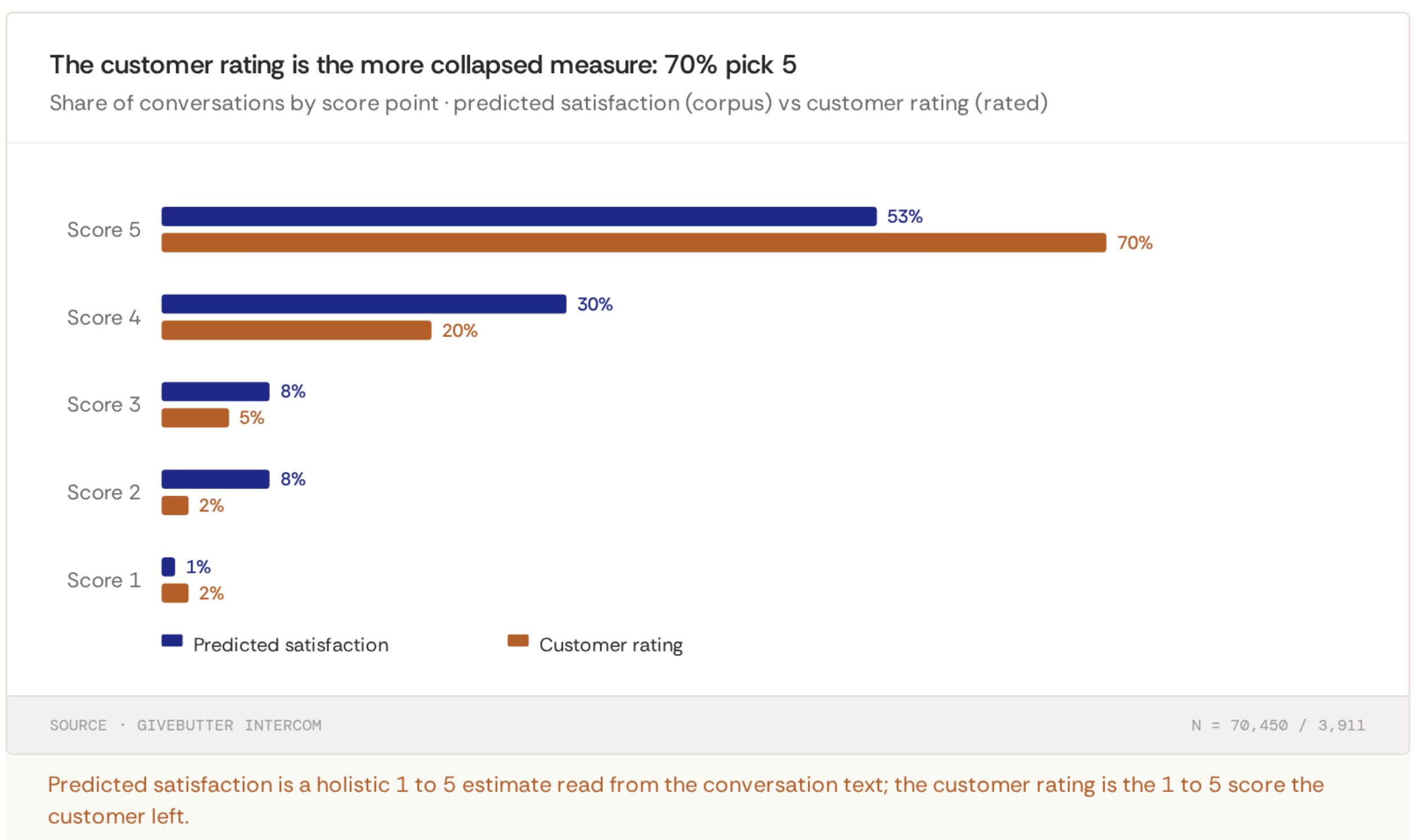

*Figure 2.*

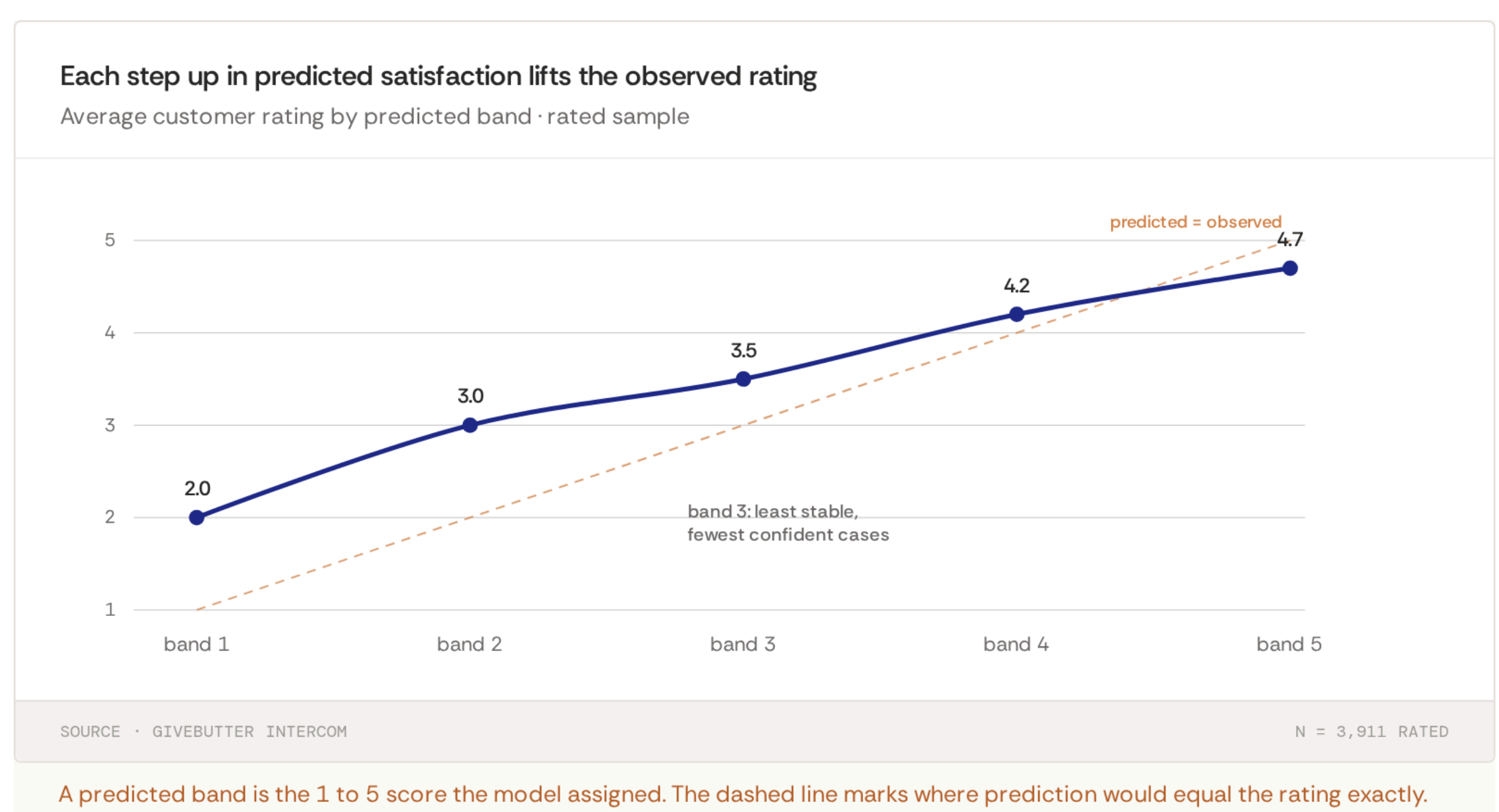

*Figure 3.*

**Table 3. Predicted-satisfaction bands map monotonically onto observed ratings.** *Rated criterion sample. The midpoint (3) is the least stable band and concentrates low-confidence cases.*

| `g2_predicted_satisfaction` | Avg. `rating_value` | Low-rating share |
|---|---|---|
| 5 | 4.7 | ~3% |
| 4 | ~4.2 | ~20% |
| 3 | ~3.4–3.6 | ~35–47% |
| 2 | ~3.0 | ~54% |
| 1 | 2.0 | ~86% |

The validation, then, settles a narrow question: between the two readings, predicted satisfaction is the one that tracks the customer's verdict, and it does so cleanly enough to use. But it also opens a wider one. If the structured estimate and the tone label were close substitutes, the better of the two would simply be the one to adopt and there would be little more to say. The rest of this paper is about what happens when they are not close substitutes at all.

## 5. Where tone and state disagree

A measure can beat its rival on a criterion and still agree with it most of the time; the winner is just right more often in the cases where they split. But that is not the case here. Across the analyzable corpus, the two readings of the same conversation disagree 44% of the time. That rate is too high to support the common working assumption that a sentiment dashboard is a rough satisfaction dashboard. If tone and state agreed on the underlying customer and merely labeled it differently, a structured satisfaction estimate would be a refinement of sentiment. A disagreement rate near one in two says they are not labeling the same thing.

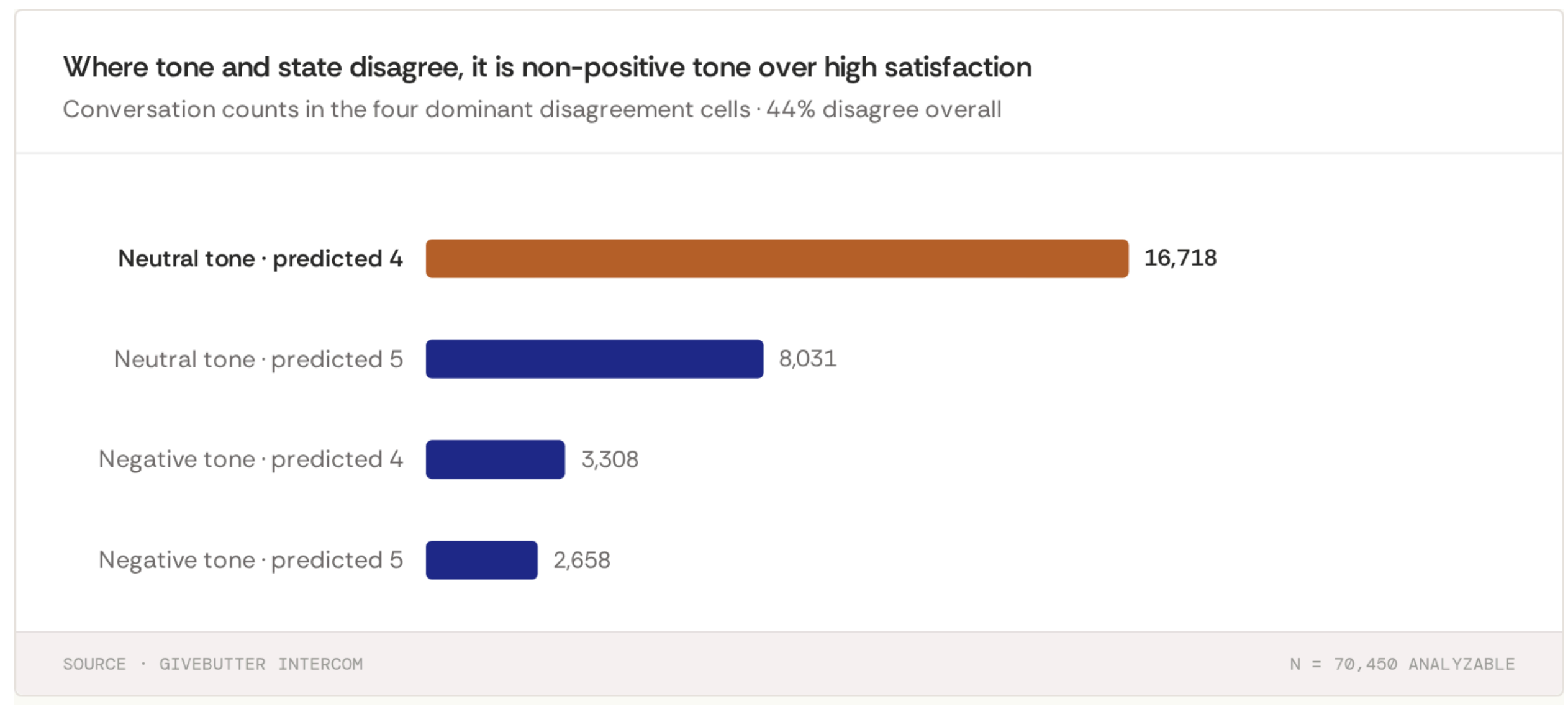

*Figure 4.*

**The disagreement has a direction.** The split is not symmetric noise. It concentrates in one quadrant: customers whose tone reads as Neutral or Negative but whose predicted satisfaction is high. The largest single cell is Neutral tone paired with a predicted 4 (16,718 conversations), followed by Neutral tone with a predicted 5 at 8,031. On the Negative side, 3,308 conversations pair negative tone with a predicted 4 and 2,658 pair it with a predicted 5. These are not edge cases scattered around the margin. They are the dominant form the disagreement takes: text that does not sound positive describing an interaction the model reads as having gone well. The mirror-image cell, positive tone with low predicted satisfaction, is comparatively rare. Whatever is driving the gap, it pushes tone below state far more often than above it.

**Table 4. Sentiment and predicted satisfaction disagree most where non-positive tone meets high predicted satisfaction.** *Analyzable corpus, N = 70,450. Cells show conversation counts for the dominant disagreement region.*

| Sentiment | Predicted satisfaction 4 | Predicted satisfaction 5 |
|---|---|---|
| Negative | 3,308 | 2,658 |
| Neutral | 16,718 | 8,031 |

**Issue presence is not negative tone in disguise.** A natural objection is that the structured measures simply repackage sentiment: that an issue report is just a negative conversation by another name. The data rule this out. Issue presence cuts across tone: 20,895 conversations pair positive tone with a reported issue, 19,480 pair neutral tone with one, and 13,485 pair negative tone with one. Customers report concrete problems while sounding positive, neutral, and negative in turn. A great many describe something broken in matter-of-fact or even friendly language. Tone does not tell you whether there is a problem to fix, and the issue flag does not tell you how the customer feels about it. They are different questions, and treating a negative sentiment label as a proxy for "something went wrong" misses most of the conversations where something did.

**The gap is not a low-confidence artifact.** The remaining objection is that disagreement is just where the model was unsure: that the structured measure wobbles on thin text and would line up with tone if we restricted to confident cases. It would not. Disagreement is actually highest in the middle, where conversations are long enough to read but not long enough to resolve: it runs to 52% for six-to-ten-message conversations and 48% for the shortest, then falls to about 36% for the longest exchanges. More tellingly, when we restrict to the cases where each measure is most certain, the disagreement persists. Conversations where predicted satisfaction is High-confidence still disagree with sentiment 45% of the time, and sentiment's own medium-confidence conversations disagree with predicted satisfaction at 66%. Confidence sharpens each reading; it does not reconcile them. Two measures that remain this far apart even at their most certain are two real signals rather than one noisy one, and the next two sections are about what each one is holding onto that the other lets go.

## 6. What the structured representation recovers

Consider the conversation that sounds fine and contains a problem. A customer writes in about an accidental tip on a donation, the agent explains the refund path, the customer says thanks, and the exchange closes politely. Sentiment files this as Positive or Neutral and moves on. But a problem was reported and resolved, and whether it was resolved well is a separate fact from how the customer sounded. Sentiment cannot hold both facts at once, because it only has one axis. The structured representation has two (was the customer satisfied, and was a concrete issue raised), and crossing them turns a single tone label into a four-cell map of operational states. The four cells are not equal in size or in meaning, and their differences are exactly what a sentiment label cannot see.

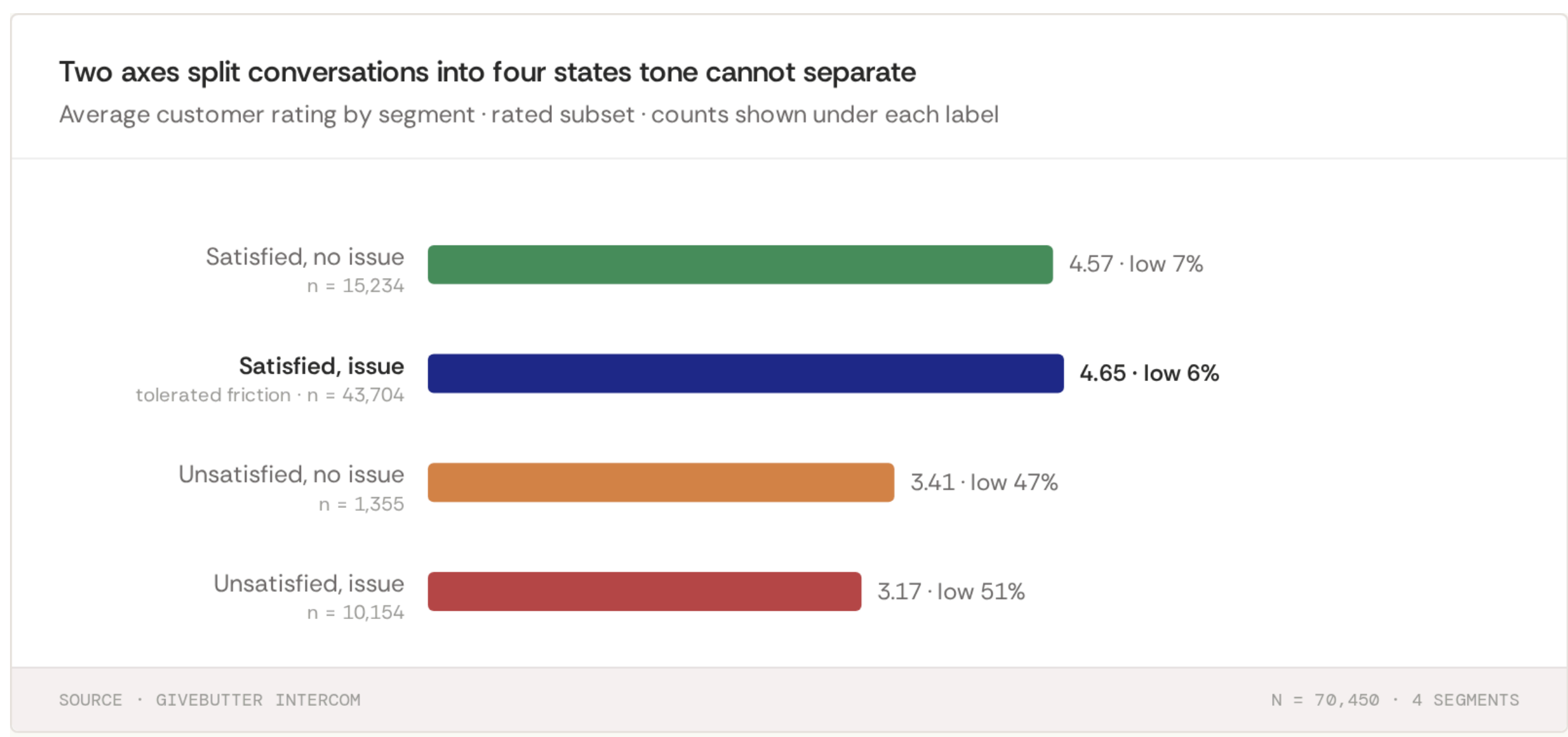

*Figure 5.*

**Four states, four very different rating profiles.** Collapsing predicted satisfaction to satisfied-versus-unsatisfied and crossing it with issue presence yields four segments with sharply different customer-rating profiles. The two unsatisfied segments rate far below the two satisfied ones (average ratings of 3.2 and 3.4 against 4.6 and 4.7), and they carry low-rating shares near 50% against single digits for the satisfied segments. A sentiment label does not separate these states; a customer in any of the four can sound neutral. The structured representation separates them cleanly, and it does so on the criterion the customer supplied, not on the model's own say-so. The implication is operational: a customer in any of these states can sound neutral, so only a read of state and cause can route each to the response it needs, where a tone dashboard sees one bucket. The state that gains most from the second axis is also the largest, the satisfied customer who still reported a problem, which the next sections take up.

**Table 5. The four structured segments separate conversations into managerially distinct states.** *Analyzable corpus. Average rating and low-rating share are computed on the rated subset within each segment.*

| Segment | Conversations | Avg. rating | Low-rating share |
|---|---|---|---|
| satisfied / no issue | 15,234 | 4.57 | 7% |
| satisfied / issue | 43,704 | 4.65 | 6% |
| unsatisfied / no issue | 1,355 | 3.41 | 47% |
| unsatisfied / issue | 10,154 | 3.17 | 51% |

**Issue detection is the axis that turns a reading into an action.** That issue presence barely moves the rating once satisfaction is known is not a weakness of the issue flag; it is the clearest evidence that the two axes measure different things. Satisfaction estimates the customer's state, and issue detection identifies the operational cause. A second axis that merely re-predicted the rating would be redundant, and a second axis that is orthogonal to it is what lets the representation say not only whether a conversation contains a risk signal but what to do about it. Issue detection is also independent of tone, cutting across Positive, Neutral, and Negative conversations alike (Section 5), so it recovers problems a sentiment reader watching only for negative affect never sees: nearly 21,000 conversations pair a reported issue with positive tone. On its own axis, issue detection is the field that converts a monitoring signal into a worklist, and it arrives with the customer's own description of what went wrong (`g2_issue_reason`) attached, so the unit of action is a concrete problem rather than a mood. Read against the literature, this is the structured-opinion view of Section 2 (Hu and Liu, 2004; Pontiki et al., 2014) applied to support: an opinion is not one polarity but a set of separable elements, and state and cause are two of them, measured here on independent axes.

**Tolerated friction is the modal conversation, and only the second axis can see it.** The single biggest segment, at 43,704 conversations, is satisfied-with-an-issue: the customer raised a concrete problem and still came away satisfied. We call this **tolerated friction**, and it is the highest-value thing the issue axis recovers. It is not a customer-experience emergency, which is precisely why leaving it unmeasured is costly. It does not trip a dissatisfaction alarm, it does not sound negative, and under any tone-based view it disappears into a positive bucket. Yet it is where fixable product and workflow problems accumulate at scale, and the issue axis surfaces it as a large, standing inventory of named friction that satisfied customers were willing to absorb but should not have had to. For a product or operations team this is the most actionable segment in the data, because each conversation already carries the specific thing to fix. The smaller mirror state, unsatisfied-with-no-issue (1,355 conversations, a 47% low-rating share), shows the same axis working in reverse: it isolates dissatisfaction with no concrete handle, the cases that need a different response precisely because there is no specific problem to point to.

> ***Friction the customer was willing to absorb.*** *"I need help setting up my fundraising account. I have 2 campaigns on my page that are the same, how do I delete one? How do I access the QR code? Where is the sharing tab?" The customer rated the conversation a 5: "Great feedback for our fundraising efforts!" (Predicted satisfaction 5, issue present.)*

**Neutral is not one state; it is several.** The clearest demonstration that tone compresses operational reality is the Neutral label itself, the modal sentiment category, which covers 28,276 conversations. Inside it, the structured segments fan out across the same range as the full corpus. Neutral-and-satisfied conversations average a rating of 4.3; Neutral-and-unsatisfied-with-an-issue conversations average 3.3, with a low-rating share of 46%. A single tone label is doing the work of describing customers who were quietly fine and customers who were quietly giving up, and it gives a manager no way to tell them apart. The issue content sharpens the contrast: within Neutral, the satisfied-with-issue conversations cluster on manageable friction like accidental tips and refund requests, while the unsatisfied-with-issue conversations shift toward unresolved support friction, account access, and inability to complete a task. Tone reports that all of these customers were unemotional. The structured representation reports what was actually happening to them.

**The value at risk is mostly unrated.** Because the validated score extends to conversations that were never rated, it can quantify exposure that rating-based monitoring cannot see. The rated subset is 5.55% of the corpus, so a team watching only customer-provided ratings is watching one conversation in eighteen. Weighting the flagged-but-unrated conversations by the fundraising volume processed behind their accounts makes the blind spot concrete: on the order of $1.5 to $2 billion in processed volume sits behind unrated conversations that are predicted-satisfied but carry an issue, and several hundred million dollars more sits behind unrated conversations that are predicted-low with an issue. These are associational, order-of-magnitude estimates over covered accounts, not precise totals, but the order of magnitude is the point. The economically meaningful exposure is not in the rated 5%. It is in the unrated majority, and only a measure that runs on every conversation can surface it.

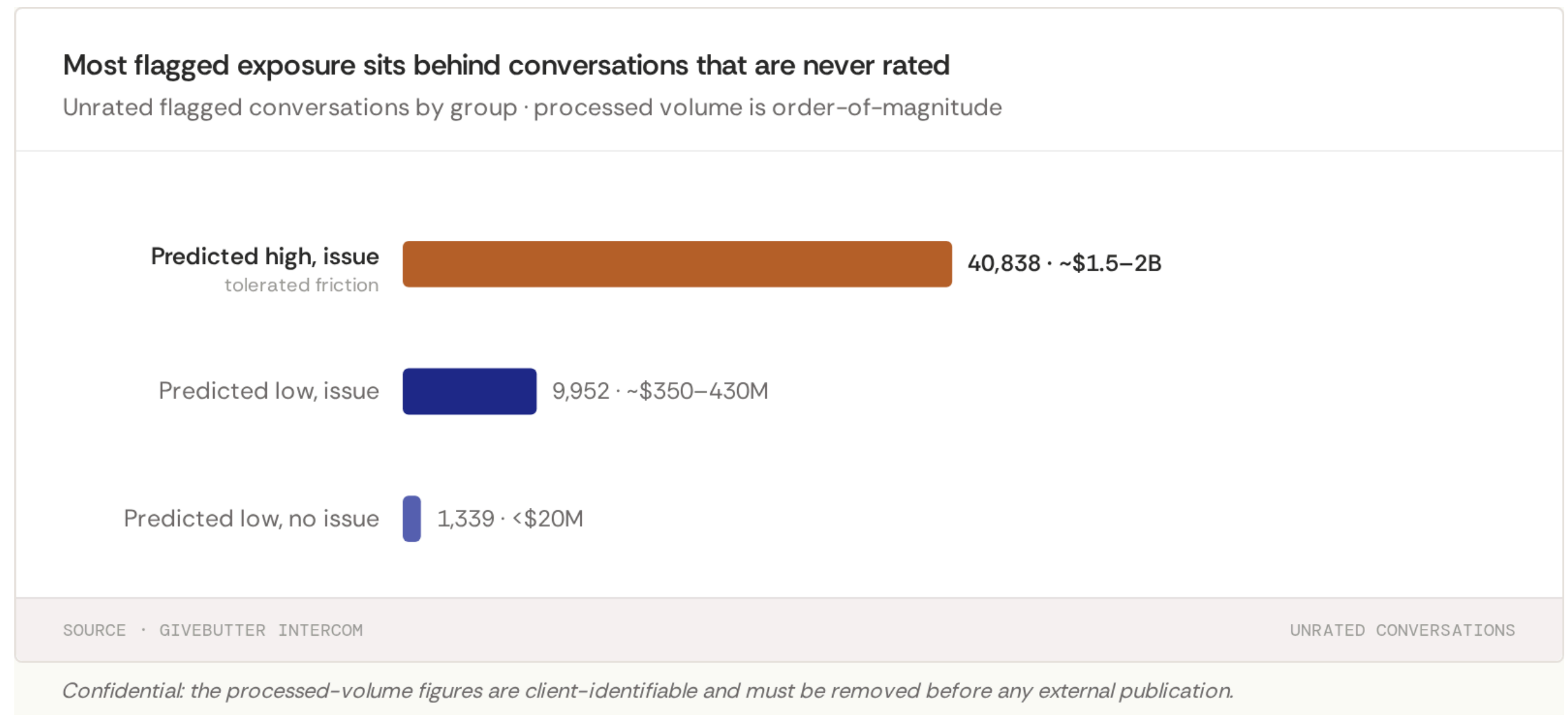

*Figure 6.*

**Table 6. Most flagged exposure sits behind conversations that were never rated.** *Unrated conversations only. Processed volume is an associational, order-of-magnitude estimate over covered accounts. Confidential: these dollar and campaign figures must be removed before publication (see editorial note).*

| Structured group | Unrated conversations | Processed volume | Campaigns |
|---|---|---|---|
| predicted high + issue | 40,838 | ~$1.5–2 billion | ~510,000 |
| predicted low + issue | 9,952 | ~$350–430 million | ~120,000 |
| predicted low + no issue | 1,339 | under $20 million | n/a |

**The advantage is broad, and it tracks independent account signals.** The structured measure's edge over sentiment is not confined to one corner of the data. Across organization verticals, channels, and tenure bands with enough rated coverage to compare, predicted satisfaction beats sentiment on criterion validity, with the largest margins in arts and culture, religious, and education organizations, in email and in-app conversations, and among newer accounts (correlation gaps from roughly +0.11 to +0.19). The advantage is broad rather than universal: in the health vertical, sentiment slightly edged predicted satisfaction on the rated subset. The structured segments also line up with account characteristics no text reading was given. The unsatisfied-no-issue segment has the lowest paid-subscription rate and the lowest verification rate of the four; the unsatisfied-with-issue segment has the highest average number of campaigns, which means dissatisfaction is not confined to small or disengaged accounts. We report these as associations and detail them, along with a noisier post-conversation disengagement analysis, in Appendix B; the timing fields available for the disengagement check were not clean enough to support a strong claim.

## 7. Why tone and state diverge: salience versus verdict

Tone and the rating diverge because they answer different questions: sentiment reads the language a customer chose, while the rating judges how the whole interaction went. A customer can be unfailingly polite about an experience they would not repeat: three civil messages about a refund that will not go through, an "ok, thanks," then a rating of 1 with the remark "too convoluted." Read as tone it is unremarkable; read as a verdict it failed. Predicted satisfaction sits closer to the verdict because it was asked to integrate the interaction, not to classify its tone.

**Salience versus verdict.** The cognitive account from Section 2 explains the gap. Open-ended writing and an overall rating run on different processes: when people describe an experience, the most accessible details dominate, the concrete and recent friction (Schwarz, 1999), whereas an overall rating is a holistic judgment weighted toward global affect (Schwarz and Clore, 1983). Sentiment reads the words, so it inherits what was salient; predicted satisfaction asks for an integrated judgment, so it approximates the verdict. The 44% disagreement is the footprint of that difference, not a defect in either reading, and it is the same divergence we found between fans' survey text and their ratings (Hong, Potteiger, and Zapata, 2026), now between two readings of one conversation rather than between a text and a separate score.

**Masked dissatisfaction and its mirror.** The mechanism predicts the two failure modes we observe, and the qualitative cases bear them out. In one direction, textual politeness and sustained engagement mask dissatisfaction: a customer cooperates throughout, the model reads the cooperation as mild satisfaction, and the rating reveals the experience was exhausting. The refund conversation rated 1 as "too convoluted," the duplicate-charge dispute rated 1 as "a definite run around" despite a courteous exchange, the setup reply rated 1 as "did not understand the question": these are **masked dissatisfaction**, where engagement is mistaken for contentment. In the other direction, tone is simply too coarse to register a poor outcome stated plainly: conversations that read as Neutral but whose remarks say "I gave up," "no help has been provided as yet," or "I didn't find Tools." Here the customer's words were informational, so sentiment recorded no affect, but the verdict was unambiguous. Both failure modes are the same construct gap seen from opposite sides, tone tracking the language while the rating tracks the outcome.

> ***Politeness is not resolution.*** *"My donation was taken out twice. I asked for a refund and sent proof. I am getting no refund. A definite run around." (Predicted satisfaction 4, customer rating 1, neutral tone.)*

**Table 7. Two readings of the same conversation, and the customer's verdict.** *Cases from the curated transcript set; internal identifiers removed, wording unchanged. In every row the model read the conversation as satisfied or neutral while the customer's own rating was the lowest score.*

| Pattern | Pred / rating | What the customer wrote | Tone recorded |
|---|---|---|---|
| Masked dissatisfaction | 4 / 1 | "This is too convoluted. Why can't I just make a refund from the transactions page?" | Neutral |
| Masked dissatisfaction | 4 / 1 | "My donation was taken out twice... A definite run around." | Neutral |
| Masked dissatisfaction | 4 / 1 | "Did not understand the question" | Neutral |
| Tone too coarse | 3 / 1 | "I didn't figure it out. I gave up" | Neutral |
| Tone too coarse | 3 / 1 | "no help has been provided as yet" | Neutral |
| Tone too coarse | 3 / 1 | "I didn't find Tools" | Neutral |

## 8. Conclusions

A sentiment label is the wrong instrument for customer support, and the cost is not subtle. A structured read of the customer's state and the operational cause recovers what a single tone label compresses away: across 70,450 conversations the two readings disagree 44% of the time, a single Neutral label stretches across average ratings from 4.3 down to 3.3, and the largest single state in the data is tolerated friction, the customer who is satisfied and has still reported a fixable problem. This rests on a validation, not an assertion. Predicted satisfaction earned the right to stand in for the customer's verdict by tracking customers' own ratings more closely than sentiment did, 0.47 against 0.36 and with roughly double the precision at the low end of the scale, where it matters most.

The reason the two readings come apart is that they answer different questions. Sentiment reads the language a customer chose, which is governed by whatever was salient enough to write down; the rating is a verdict, an integrated judgment that leans on affect and the shape of the experience rather than an inventory of its parts. A structured representation that estimates the customer's state and flags the operational cause sits closer to the verdict and, unlike tone, carries the information a team needs to act. It separates tolerated friction from acute dissatisfaction, splits the uninformative Neutral bucket into states that range from quietly fine to quietly giving up, and, because it runs on every conversation rather than the rated minority, surfaces exposure that rating-based monitoring never sees.

Ultimately how a customer sounds is not the same as whether they were helped, and the measure of satisfaction and tone don’t strictly match in almost half of all conversations. Reading whether a customer was satisfied, and what problem they ran into, tracks how they actually rated their support better than sentiment. Companies relying on sentiment analysis will miss the satisfied customer who still hit a fixable problem, the unhappy customer who stayed polite, the "Neutral" label that hides both, and the risk buried in the conversations no one ever rated. Prior to the availability of large language models measuring how customers sound was the best option because it was the only option. Today the enterprise has the ability to extract multiple dimensions of meaning from individual records that are more precise and flexible than sentiment. Emerging research methods that employ predictive scoring using LLMs are becoming increasingly attractive due to their increased flexibility and scalability.

## 9. Implications for data practitioners

**Monitor state and cause, not tone.** A sentiment dashboard is not a satisfaction dashboard. The two disagree 44% of the time here, often enough that using one to stand in for the other will mislead you. Track two things instead: whether the customer was satisfied (predicted satisfaction) and whether they reported a problem (issue presence). Tone is still worth watching, but it should not be what drives what you prioritize, escalate, or dig into.

**Use predicted satisfaction as your main signal, and trust its precision more than its recall.** It lines up with customers' own ratings better than sentiment does, and its real strength is precision: when it flags a conversation as low, it is right about half the time, versus about a quarter for sentiment. For a review queue, that hit rate is the whole game. Read the score as a risk level: a 1 or 2 means look now, a 3 is a coin flip worth a glance, a 4 is low risk but worth a check for issues, and a 5 is safe to skip. Because the score is calibrated against real ratings, you can run it as a live rule on the 95% of conversations that never get rated, instead of only looking back after the fact.

**Treat tolerated friction as its own category, and treat it as an asset.** The biggest group in the data is the customer who was satisfied but still reported a problem, and it is the most useful thing the issue flag gives you. Do not let it vanish into the happy pile. It will not trip a dissatisfaction alert and it is not an escalation, but it is where fixable product and workflow problems pile up, and each conversation comes with the customer telling you exactly what to fix. That is tens of thousands of conversations here, and behind the unrated ones, fundraising volume in the billions. Send it to product and operations, not to customer recovery; track it apart from your urgent unhappy customers; and mine those problem descriptions for the small fixes that quietly lift satisfaction across the whole customer base.

**Distrust the Neutral bucket most of all.** If one category deserves a second look, it is Neutral sentiment: it is the most common label and the least useful. It holds customers who were quietly fine and customers who were quietly giving up, with average ratings running from 4.3 down to 3.3. Any workflow that reads Neutral as "nothing to see here" is throwing away the group most likely to hide a bad outcome. Before you act on it, split it by whether the customer was satisfied and whether they reported a problem.

## 10. Limitations

**The criterion is limited and skews positive.** Customer ratings exist for 5.55% of conversations and are heavily concentrated at the top of the scale, which inflates raw agreement and makes estimates least stable exactly where they matter most, at the low end. We mitigated this by leading with skew-robust metrics and reporting collapsed agreement against the majority-class baseline, but the low-rating cells are small (355 low ratings) and the calibration of the bottom bands should be read as directional. The favorable skew also cannot be fully separated from the satisfaction function's conservative prior, as Section 7 notes; that prior is itself consistent with the documented bunching of LLM scalar scores (Licht, 2025), a known measurement risk the criterion validation only partly addresses.

**The raters are not the corpus.** Customers who leave a rating differ from those who do not: they have slightly longer conversations, are somewhat more positive on both tone and predicted satisfaction, are more email-heavy, and sit at slightly smaller accounts. Validation metrics computed on the rated subset therefore may not transport unchanged to the full population. The direction of the bias is toward a more favorable, more engaged subset, and the validation advantage is large enough that modest selection effects are unlikely to reverse it, but the population effect size may differ from the rated-sample estimate.

**Account outcomes are associational and snapshot-timed.** The independent account signals in Section 6 (subscription, verification, campaign counts, value at risk) are observed at the account level at a single point in time. They license association, not causation. The post-conversation disengagement check was weaker still: many activity timestamps predate the conversation, so the timing is not cleanly forward-looking, and we treat those results as suggestive only.

**The annotations are model-derived, and the issue taxonomy is dynamic.** Predicted satisfaction and issue presence are estimates from a single model run; the claim is not that they are perfect but that they are more useful than tone as an operational representation. The issue-reason field is a free-text label that requires downstream consolidation before its categories are comparable, so the issue-content patterns in Section 6 should be read as indicative rather than as a validated taxonomy.

**One model, one domain, one window.** The annotations come from a single model (GPT-5.4) over a single six-month window, in one domain: nonprofit fundraising software support. The construct gap between tone and state is likely general (its mechanism is cognitive, not domain-specific), but the magnitudes here are evidence within this setting, not a measurement that transports unchanged to other industries, models, or periods.

## Appendix A. Coding schema (prompt functions)

All annotations were produced by independent extract_session_info prompt functions run with GPT-5.4 on the Dimension Labs platform. Each function reads one conversation's text and returns the fields below. Detail fields in the issue function are null-gated on g2_issue_mention.

### A.1 User sentiment function

| Field | Type | Values / definition |
|---|---|---|
| `g2_user_sentiment` | enum | Positive / Neutral / Negative tone of the customer's messages. |
| `g2_user_sentiment_confidence` | enum | High / Medium / Low confidence in the sentiment label. |

### A.2 Issue mention function (gate-and-detail)

| Field | Type | Values / definition |
|---|---|---|
| `g2_issue_mention` | boolean | Whether the customer reports a specific problem (gate). |
| `g2_issue_mention_confidence` | enum | High / Medium / Low; null when gate is False. |
| `g2_issue_reason` | string | Short (<7 words) summary of the issue; null when gate is False. |
| `g2_issue_verbatim` | string | Customer wording for the issue (<20 words); null when gate is False. |

### A.3 Predicted satisfaction function

| Field | Type | Values / definition |
|---|---|---|
| `g2_predicted_satisfaction` | integer | Holistic 1–5 estimate (1 Very Dissatisfied … 5 Very Satisfied). |
| `g2_predicted_satisfaction_confidence` | enum | High / Medium / Low; how diagnostic the text is for the assigned point. |
| `g2_predicted_satisfaction_rationale` | string | Reasoning trace for the score (<15 words). |

### A.4 Context function (corpus gates and handling)

| Field | Type | Values / definition |
|---|---|---|
| `g2_conversation_true` | boolean | A genuine two-party conversation occurred. |
| `g2_full_conversation_true` | boolean | The conversation is complete, not a fragment. |
| `g2_message_count` | integer | Number of messages; gate requires >= 3. |
| `g2_rating_request` | boolean | Whether a rating was solicited in the interaction. |
| `g2_agent_sentiment` | enum | Tone of the support agent's messages. |

| Field | Type | Values / definition |
| --- | --- | --- |
| g2_success_rating | integer | 0–10 estimate of how well the agent handled the request. |

## Appendix B. Supplementary tables

**B.1 Segment moderation.** The predicted-satisfaction advantage over sentiment (correlation gap against the customer rating) by segment, among segments with enough rated coverage to compare. The advantage is broad but not universal.

| Segment | Correlation gap (predicted − sentiment) |
|---|---|
| Arts, culture, and humanities | +0.162 |
| Religion-related | +0.140 |
| Education | +0.138 |
| Email channel | +0.121 |
| In-app channel | +0.113 |
| Newer-tenure accounts | +0.115 to +0.188 |
| Health (sentiment higher) | −0.029 |

**B.2 Account-outcome associations.** These are associational, not causal. The unsatisfied / no-issue segment has the lowest paid-subscription rate (12.8%) and the lowest verification rate (45.5%) of the four segments; the other three cluster near 19–20% subscription. The unsatisfied / issue segment has the highest average number of campaigns (19.45), indicating that dissatisfaction is not confined to small or low-engagement accounts.

**B.3 Post-conversation inactivity (heuristic, suggestive only).** Many activity timestamps predate the conversation, so the timing is not cleanly forward-looking; these rates should not carry a strong claim.

| Structured segment | Inactivity rate |
|---|---|
| satisfied / issue | 56% |
| satisfied / no issue | 61% |
| unsatisfied / issue | 62% |
| unsatisfied / no issue | 89% |

**B.4 Selection and representativeness.** Customers who leave a rating differ from those who do not: they have slightly longer conversations, are somewhat more positive on both tone and predicted satisfaction, are more email-heavy, and sit at slightly smaller accounts. Validation metrics computed on the rated subset may therefore not transport unchanged to the full population, though the direction of the bias (toward a more favorable, more engaged subset) and the size of the advantage make a reversal unlikely.

## Appendix C. Verbatim reference examples

Cases drawn from the curated transcript set, organized by divergence pattern. Internal conversation identifiers are removed and customer wording is unchanged; the language quoted is the customer's own, including blunt phrasing. Pred / rating is the model's predicted satisfaction and the customer's own 1–5 rating. These cases are illustrative, not prevalence estimates.

**C.1 Masked dissatisfaction.** Predicted satisfied, rated at the floor: the model read cooperation as contentment while the customer's verdict was a 1.

| Tone | Pred → rating | What the customer wrote | Issue |
|---|---|---|---|
| Neutral | 4 → 1 | "This is too convoluted. Why can't I just make a refund from the transactions page?" | ticket refund |
| Neutral | 4 → 1 | "My donation was taken out twice... A definite run around." | duplicate charge |
| Neutral | 4 → 1 | "Did not understand the question" | can't start over |

**C.2 Tone too coarse.** Neutral wording, failed outcome: the language carried no affect, so sentiment recorded none, but the task was never completed.

| Tone | Pred → rating | What the customer wrote | Issue |
|---|---|---|---|
| Neutral | 3 → 1 | "I didn't figure it out. I gave up" | can't find auction tab |
| Neutral | 3 → 1 | "no help has been provided as yet" | manage recurring donation |
| Neutral | 3 → 1 | "I didn't find Tools" (abandoned) | can't find Tools tab |

**C.3 Overt dissatisfaction.** Here tone and verdict agree; these anchor the unsatisfied-with-issue cell, the clearest low-rating state.

| Tone | Pred → rating | What the customer wrote | Issue |
|---|---|---|---|
| Negative | 1 → 1 | "never resolved, wasted my time..." | DNS not verifying |
| Negative | 2 → 1 | "Your website sucks... The entire system is horrific" | can't log in to manage donation |
| Negative | 2 → 1 | "answer was wrong! should have sent me to a human faster" | live auction bids not shown |
| Negative | 3 → 1 | "the information does not resolve my issue" | can't edit payout info |

**C.4 Issue without dissatisfaction.** The issue flag fires even when the customer is satisfied, which is why issue presence and satisfaction are distinct axes.

| Tone | Pred → rating | What the customer wrote | Issue |
|---|---|---|---|
| Neutral | 5 → 5 | "Great feedback for our fundraising efforts!" (with a stack of setup questions) | campaign setup |
| Neutral | 3 → 4 | "Goat experience for Bangladesh" | personal campaign by location |